%% file: 0_main.tex
\documentclass[letterpaper, 10 pt, conference]{ieeeconf}  

\IEEEoverridecommandlockouts                              

\usepackage{graphicx} 
\usepackage{amsmath} 
\usepackage{mathtools}
\usepackage{amssymb}  
\usepackage{booktabs}
\usepackage{soul}

\usepackage{algorithm}
\usepackage{algpseudocode}
\usepackage{graphicx}
\usepackage{xcolor}

\usepackage{amsthm}
\usepackage{tabularx}
\usepackage{siunitx}
\usepackage{multirow}
\usepackage{flushend}

\usepackage[font=scriptsize,labelfont=bf]{caption}
\newcommand{\numberthis}[1]{\addtocounter{equation}{1}\tag{\theequation}\label{#1}}

\makeatletter
\let\NAT@parse\undefined
\makeatother
\usepackage{hyperref}  

\newcommand{\T}[3]{\prescript{#2}{}{#1}_{#3}}
\newcommand{\etal}{\emph{et al.}}
\newcommand{\ie}{\textit{i.e.}}
\newcommand{\eg}{\textit{e.g.}}

\newcommand{\smalltitle}[1]{{\noindent\textbf{#1}}}

\title{\LARGE \bf
    Model Predictive Control for Fluid Human-to-Robot Handovers
}

\author{

Wei Yang$^{\dagger *}$, Balakumar Sundaralingam$^{\dagger *}$, Chris Paxton$^{\dagger *}$, Iretiayo Akinola$^{\dagger}$, \\
Yu-Wei Chao$^{\dagger}$, Maya Cakmak$^{\dagger,\ddagger}$, Dieter Fox$^{\dagger,\ddagger}$
    \thanks{$*$ Equal contribution.}
    \thanks{$^{\dagger}$NVIDIA, USA
            {\tt\small \{weiy, balakumars, cpaxton, iakinola, ychao, mcakmak, dieterf\}@nvidia.com}}%
    \thanks{$^{\ddagger}$University of Washington, USA}
}

\begin{document}

\maketitle
\thispagestyle{empty}
\pagestyle{empty}

\begin{abstract}
\input{1_abstract}
\end{abstract}

\section{Introduction}\label{section:intro}
\input{2_introduction}

\section{Related Work}\label{section:related_work}
\input{3_related_work}

\section{Task Model}\label{section:task_model}
\input{4_task_model}

\section{Perception Pipeline}\label{section:perception}
\input{5_perception}

\section{Model Predictive Control for Grasping}\label{section:mpp}
\input{6_mppi}

\section{Systematic Evaluation}\label{section:sys_eval}
\input{7_systematic_evaluation}

\section{User Evaluation}\label{section:user_study}
\input{8_user_study}

\section{Discussion and Conclusion}\label{section:conclusion}
\input{9_conclusion}




\input{10_appendix} 




\bibliographystyle{IEEEtran}
\bibliography{handover}

\end{document}

%% file: 1_abstract.tex
Human-robot handover is a fundamental yet challenging task in human-robot interaction and collaboration.
Recently, remarkable progressions have been made in human-to-robot handovers of unknown objects by using learning-based grasp generators. 
However, how to responsively generate smooth motions to take an object from a human is still an open question.
Specifically, planning motions that take human comfort into account is not a part of the human-robot handover process in most prior works.
In this paper, we propose to generate smooth motions via an efficient model-predictive control (MPC) framework that integrates perception and complex domain-specific constraints into the optimization problem. 
We introduce a learning-based grasp reachability model to select candidate grasps which maximize the robot's manipulability, giving it more freedom to satisfy these constraints. 
Finally, we integrate a neural net force/torque classifier that detects contact events from noisy data. 
We conducted human-to-robot handover experiments on a diverse set of objects with several users ($N=4$) and performed a systematic evaluation of each module. 
The study shows that the users preferred our MPC approach over the baseline system by a large margin.

%% file: 2_introduction.tex
Recently, remarkable advances have been made in generic human-to-robot handovers of arbitrary graspable objects~\cite{rosenberger2020object,yang2021reactive} thanks to development of learning-based grasp planners for unknown objects~\cite{morrison2018closing,mousavian2019graspnet}. 
In these works, grasps are selected based on grasp stability or task constraints, and the robot is driven towards the end-effector pose to complete the handover process.

While promising results have been demonstrated using this pipeline, how to select the best grasp and associated motion to execute is still an open challenge.
Prior works select grasps based on grasp quality~\cite{rosenberger2020object,yang2021reactive}, which neither guarantees the reachability and manipulability of the grasp nor the smoothness of the motion generated to reach the grasp. 
In this paper, we propose to learn a neural network to predict the manipulability of a grasp pose for motion-aware grasp selection. 
Together with the grasp stability, it enables us to select a stable grasp which is also reachable and manipulable. 
Additionally, compared with using inverse kinematics to check the reachability of each grasp sequentially, our model can evaluate the reachability and the manipulability of a batch of grasps in parallel, which is more efficient for real-time robotics tasks such as human-to-robot handovers. 

Another challenge is how to plan smooth and natural motion for handovers. 
Given a selected grasp, the robot is usually driven towards the end-effector pose by local (non-global) policies such as Riemannian Motion Policies~\cite{ratliff2018riemannian} as in ~\cite{yang2021reactive,yang2020human} or visual servoing~\cite{rosenberger2020object}. 
However, natural motions are not just accelerations towards a particular end-effector pose; they consider how to reach there while obeying complex task constraints. 
In addition, without planned motions, the robot might choose a grasp that leads to slower or less efficient motions. 
In this work, we employ a highly-parallelized Stochastic Model Predictive Control (MPC) framework~\cite{bhardwaj2021fast} for real-time motion planning. 
This framework also allows us to incorporate a range of complex task-specific constraints proposed in~\cite{yang2020human} such as preventing hand occlusions.

\begin{figure}
    \centering
    \includegraphics[width=1\linewidth]{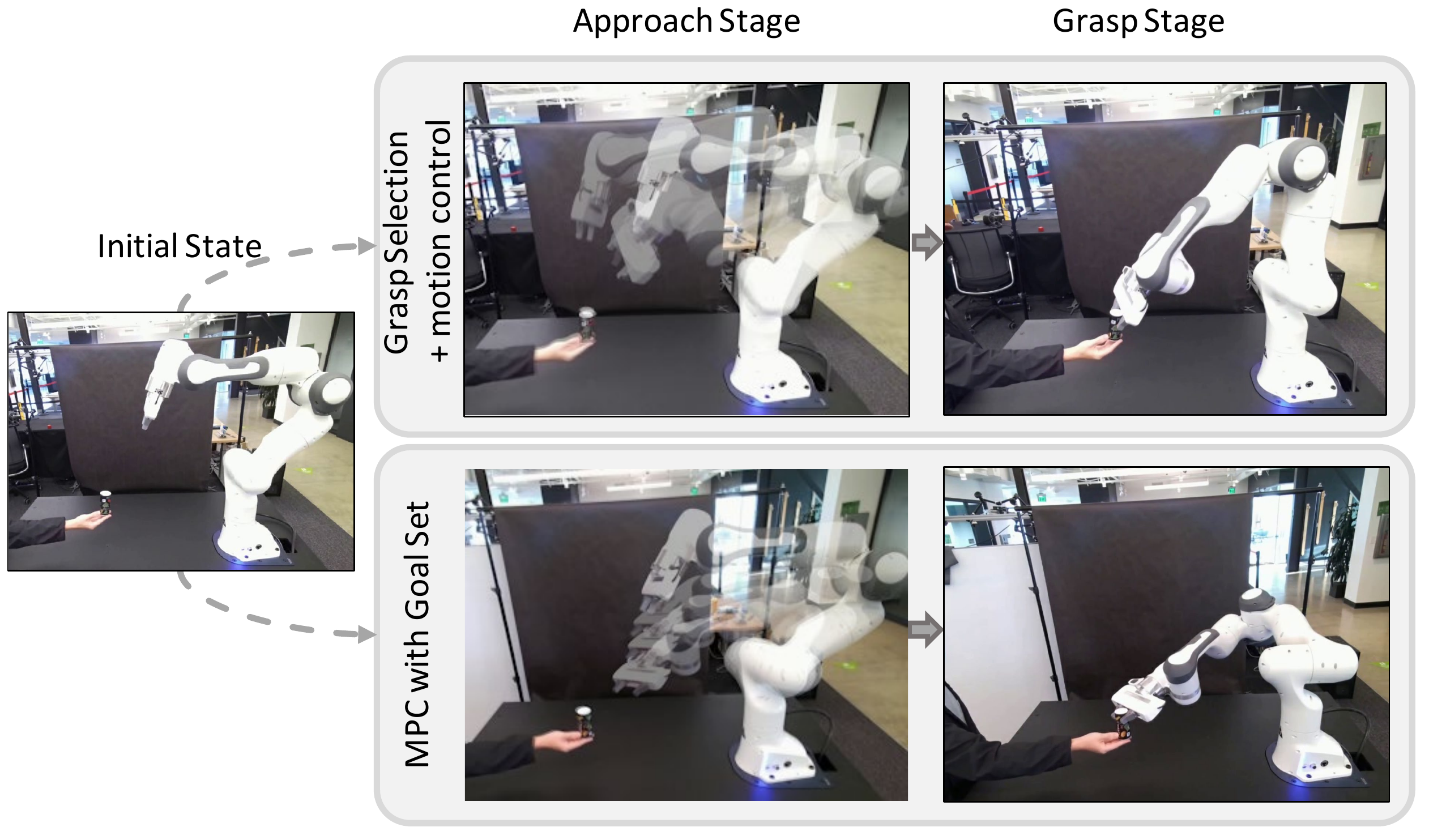}
    \caption{
    How to responsively generate smooth motions for handovers is an open question. 
    The pipeline of planning motion after selecting a grasp, such as in~\cite{yang2021reactive}, does not always guarantee smooth motions (top). 
    We propose an approach to use Model-Predictive Control (MPC) for motion generation with either a single selected grasp or a set of grasps (bottom), which results in a better, more fluid user experience. 
    More results and videos are available at \url{https://sites.google.com/nvidia.com/mpc-for-handover}.
    }
    \label{fig:teaser}
    \vskip -18pt
\end{figure}

To bridge the gap between grasp selection and motion planning, we further extend the MPC framework by incorporating the grasp selection procedure into the optimization given a set of candidate grasps. 
This unified MPC framework enables the system to select the grasp and plan motion simultaneously, while taking grasp reachability and domain specific constraints into account. 
\begin{figure*}[htp]
    \centering
    \includegraphics[width=0.9\textwidth]{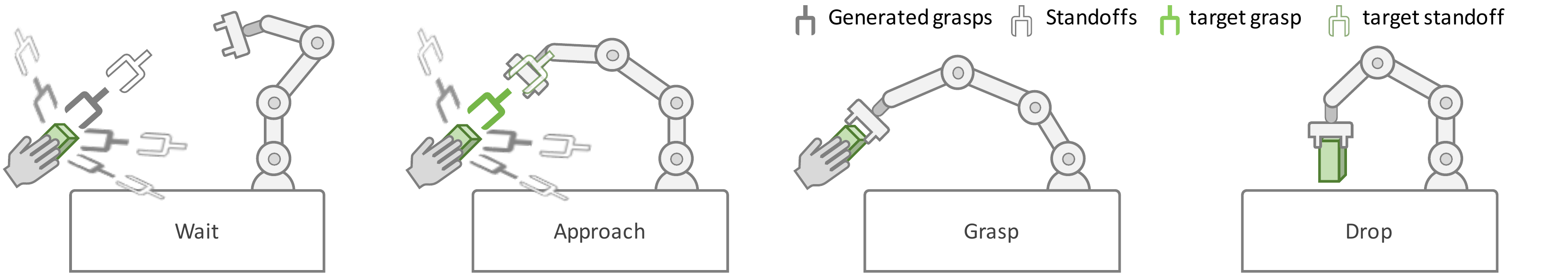}
    \caption{Different stages of human-to-robot handovers. 
    The robot is idle until it detects a human holding an object (\textit{Wait}). 
    Then it approaches to a standoff position using MPC (\textit{Approach}). 
    Once the robot reaches the standoff pose, the robot grasps the object with a blocking policy (\textit{Grasp}).
    Finally, the robot places the object to a preset position (\textit{Drop}). 
    }
    \label{fig:task_model}
    \vskip -12pt
\end{figure*}

Finally, during the physical handover phase, \ie, from the first contact of the receiver’s hand on the object till the release of the object, force feedback can be used to decide when to release the object (robot-to-human) or when to grasp the object (human-to-robot)~\cite{costanzo2021handover}. 
However, the raw force/torque sensor readings are often noisy and hard to interpret directly~\cite{chan2012grip}. 
To handle this, we collected a dataset of humans interacting with a moving robot, and trained a neural net to detect when the contact between hand and gripper happens. 
The detection is then coupled with vision during the physical handover phase to trigger the robot grasp action. 

Our contributions are summarized as follows: 
1) We integrate a stochastic MPC framework for efficient motion planning with complex task-specific objectives, and proposed a unified framework for grasp selection and motion planning. 
2) We introduce a learned reachability/manipulability model to facilitate motion-aware grasp selection. 
3) We collected a dataset and trained a neural net to detect the contact events for improving action triggering in the physical handover phase. 
4) We evaluate handovers with the proposed systems on three objects from different handover locations and orientations, as well as dynamically changing the object orientation after the robot starts to move.  
We further conducted a user evaluation ($N=4$) with a diverse set of household objects comparing the proposed system with the baseline.

%% file: 3_related_work.tex
Human-robot object handover is essential for assistive and collaborative robots in environments such as retirement homes and factories. Many efforts have been made in human-robot communication, grasp planning, perception, motion planning, and control to enable natural and fluent handovers~\cite{ortenzi2020object}. 
Among them, grasp planning is a crucial component in pre-handover phase to ensure a success physical handover. 
For example, human gaze was used to determine the contact point on the object to direct a grasp in~\cite{johansson2001eye}. 
Cini \etal~\cite{cini2019choice} chose grasps based on the object constraints and the receiver's task. 
For human-to-robot handovers, object properties such as shape, function and safety were considered when planning grasp in~\cite{kim2004advanced}. 
Yang \etal~\cite{yang2020human} decided robot grasps based on how the object is holding by a human giver.
Recent advances~\cite{rosenberger2020object, yang2021reactive} used deep-learning based grasp planners to generate grasps for unknown objects in human hand.
These works selected the grasp to be executed based on the grasp quality or the stability associated with the grasps, which could not guarantee the selected grasps are all manipulable. 
Our reachability model directly evaluates the robot reachability/manipulability given a grasp and is complementary to the grasp stability and quality metrics. 

Model predictive control~(MPC) is being increasingly used for reactive motion generation in manipulators~\cite{bhardwaj2021fast, faroni2019mpc,minniti2021model,kramer2020model, kuindersma2016optimization}. Kuindersma~\etal~\cite{kuindersma2016optimization} leveraged MPC to generate dynamic motions for humanoid robots, that can navigate varying terrains. Faroni~\etal~\cite{faroni2019mpc} explored leveraging MPC to dynamically slow down the robot when a human is in the workspace of the robot. Minniti~\etal~\cite{minniti2021model} used MPC in combination with an observer to handle contact with the environment such as contact during opening of doors. Kramer~\etal~\cite{kramer2020model} showed their approach avoiding dynamic obstacles leveraging non-linear programming to solve MPC. Bhardwaj~\etal~\cite{bhardwaj2021fast} took a different approach from the above methods for MPC, specifically they explore leveraging sampling-based MPC for real-time reactive manipulator control. 
By leveraging sampling, their approach does not require cost terms to be smooth or differentiable. We leverage~\cite{bhardwaj2021fast} in this paper to solve MPC for human-robot handovers as it does not require 1) extensive engineering effort to tune the many heuristics and 2) the cost terms to be smooth/differentiable.

%% file: 4_task_model.tex
Our high-level task model largely follows the one in~\cite{yang2021reactive,yang2020human}. 
As illustrated in Fig.~\ref{fig:task_model}, there are four states the robot can be in: (1) waiting for human or object, (2) approaching the object standoff position, (3) grasping the object, and (4) placing the object when it is done.

\textbf{(1) Wait.} During this stage, the robot is idle. It will move to a home position, and not otherwise move around. This step will continue until the robot detects a human holding an object in the workspace.

\textbf{(2) Approach.}  During this phase, the robot approaches the object and reaches a standoff pose from the object. This standoff pose is 15 cm from the final grasp on the object. Getting to this standoff position is the role of the MPC system discussed in~Sec.\ref{section:mpp}. 

\textbf{(3) Grasp.} Once the robot reaches the standoff pose, we begin our grasping policy. Unlike the \emph{Approach} or \emph{Wait} policies, this is a \textit{blocking} policy because as the robot gets closer to the object, body tracking, segmentation, and grasp estimation all become less reliable, as discussed in~\cite{yang2021reactive}. After moving the end-effector forward to the grasp position, the robot closes its gripper and retreats to the standoff pose.

We detect missed grasps by observing the distance between the gripper fingers after closing. After a failure grasp, the robot goes back to the \emph{Approach} phase, and attempts to grasp the object again. 

We trained a classifier given raw force/torque signal to detect the contact with the gripper. If the human pushes the object into the robot's hand, the grasp motion will terminate early and the gripper will be closed. The force classifier is described in more detail in Sec.~\ref{sec:force}.

\textbf{(4) Drop.} If the robot has the object, it will place it on the table at a preset position.

%% file: 5_perception.tex
In this section, we first briefly review the modules we used for grasp generation in Section~\ref{sec:grasp-generation}. 
Then we present our reachability model which measures the manipulability of a grasp pose in~\ref{sec:reachability}. 
Finally, a contact event detector is introduced in~\ref{sec:force} to improve the robustness of the system when the contact happens (\eg, when humans push/pull the object into/from the robot gripper or when the robot gripper touches the object) or when vision modules fail. 

\subsection{Grasp Generation}~\label{sec:grasp-generation}
Our system generates a set of candidate grasps for human-to-robot handovers by following~\cite{yang2021reactive}. 
Specifically, we track the robot pose by DART~\cite{schmidt2014dart} and the human body by the Azure body tracking SDK at 15 Hz, then segment the hand and the object in hand by a pretrained hand segmentation model at 9 Hz. Given the segmented object point cloud, we use a temporally consistent version of 6-DoF GraspNet~\cite{yang2021reactive, mousavian2019graspnet} to generate grasps at 5 Hz. 

\subsection{Grasp Reachability Model}~\label{sec:reachability}
While GraspNet~\cite{mousavian2019graspnet} estimates a score for each predicted grasp to measure the grasp stability, not all the generated stable grasps can be realized by the robot. 
To reduce the set of grasps being considered, we introduce a learned ranking/cost function that predicts the manipulability of grasp pose without an intermediate Inverse Kinematics (IK) step.
The manipulability of a grasp pose is a measure of how far the robot is to a singular configuration \cite{vahrenkamp2012manipulability}. 
This metric is particularly valuable when trying to reach a moving target; a highly manipulable grasp is likely to stay reachable as the object moves.
While previous works have developed different methods for encoding reachability of grasps, their approaches use simple heuristics \cite{berenson2007grasp}, proxy representations \cite{akinola2018workspace} or require interpolation of precomputed data~\cite{vahrenkamp2009humanoid,akinola2021dynamic}. Our approach directly learns the manipulability metric from data to quickly predict scores. 

Concretely, our manipulability metric $\mathcal{M}$ is defined as $|J^T J|$ where $J$ is the Jacobian of the IK of the grasp $g$. 
If no IK solution exists, we use the negative twist-distance between the closest achievable end-effector pose and the target: 

\[
\mathcal{M}(g) = 
\begin{cases}
    |J^T J|, & \text{if IK exists},\\
    - \si{dist_{twist}}(g, e),              & \text{otherwise},
\end{cases}
\]
\noindent  where $|\cdot|$ is the determinant operator, $\si{dist_{twist}}(g, e)$ is the twist distance between  grasp $g$ and the closest achievable pose of the end-effector pose $e$.

We further incorporate joint limits of the robot into the manipulability score, we use a joint-limit performance metric derived in previous work~\cite{chan1995weighted} to form a weighting matrix. 
The metric for each joint $\theta = \{\theta^{i} : i = 1 ,2, \cdots , \si{dof}\}$:

\begin{align*}
\mathcal{P}_{\text{joint\_limit}}(\theta) &= \dfrac{(\theta_{max} - \theta_{min})^ 2 * (
                    2  \theta - \theta_{min} - \theta_{max})}{4  (\theta_{max} - \theta) ^ 2 * (\theta - \theta_{min}) ^ 2} \numberthis{eq:jl_weight}.
\end{align*}
Note that this metric is minimum ($0$) in the middle for the joint ranges and maximum (infinity) at the extremes of the joint limits. The weighting matrix is a diagonal matrix with entry on the diagonal corresponding to each joint given as
\begin{align*}
    \mathcal{W}^{ii} &= \dfrac{1}{\sqrt{1 + |\mathcal{P}_{\text{joint\_limit}}(\theta^{i})|}} \numberthis{eq:jl_2}.
\end{align*}
The joint-aware manipulability is then computed as $|J^T \mathcal{W} J|$ for grasp poses that have IK solutions. This is used to generate the dataset that is used to train the joint-limit-aware reachability model.

To facilitate training, we generated a dataset of grasp poses and the corresponding manipulability scores, and trained an multi-layer perception with the mean squared error loss. 
To account for IK redundancies during data generation, the manipulability of a given grasp pose is the maximum of the manipulability scores over the IK solutions that realize that grasp.
The input of the multilayer perceptron (MLP) is the 6-DOF grasp pose $(\si{x}, \si{y}, \si{z}, \si{roll}, \si{pitch}, \si{yaw}) \in \mathbb{R}^6$ and the output is the manipulability score.  
The median inference time of the trained model is 1.3 milliseconds in experiments. 

\subsection{Physical Contact Detection}~\label{sec:force}
To better coordinate the timing for the physical handover phase, when the visual perception becomes noisier due to the close proximity between the hand and the robot gripper, we trained a feed-forward neural network to detect the contact event between the hand/object and the robot gripper.
The network takes the raw sensor data including joint velocities, efforts, force, and torque for the last $T=5$ steps as input, and predicts a probability of a contact event.

Our model is a temporal convolutional neural net which encodes each individual timestep through a two-layer MLP, then performs two 1D convolutions over history, before another MLP (with dropout) predicts a single output indicating if a force event was detected within the window.

We collected a physical contact dataset mimicking the handover procedures: 
we moved the robot gripper to a random position in the workspace. When the robot was about to reach the target pose, we pushed/pulled on the robot gripper, and pressed a key to label the physical contact. We recorded joint positions, velocities, efforts, forces, and torques during the whole procedure.
With $T=5$, our dataset consists of 20K samples with about $8\%$ positive samples. 

Our model achieved $93.6\%$ accuracy on a held-out non-overlapping test set of 5k examples collected. 

%% file: 6_mppi.tex
For human-to-robot handover, the robot needs to move to one of the grasp poses~$X_{g\in G}$, and then grasp the object from the human, as shown in Fig.~\ref{fig:task_model}. During this robot motion, we also want to encode heuristics that promote fluid human-robot handovers, specifically:
1) encouraging the robot's gripper to move in straight lines;
2) avoiding collisions between robot and human; 
3) having the human hand always in view; and
4) reducing jerk of the robot during motion.
As some heuristics are in the joint space and some in Cartesian space of the links, we formulate the problem as a kinematic joint space trajectory optimization problem~\cite{sundaralingam-2019auro-relaxed,toussaint2014newton} (Sec.~\ref{sec:traj_opt}). We then solve the formulated problem in real time leveraging stochastic MPC~\cite{bhardwaj2021fast} (Sec.~\ref{sec:mpc_solve}). 

\subsection{Trajectory Optimization Formulation}
\label{sec:traj_opt}
Given the robot's joint position~$\theta_0$ and velocity~$\dot{\theta}_0$ at timestep~$0$, we want to compute the joint accelerations~$\ddot{\theta}_{t\in[0,H-1]}$ across the horizon~$H$ timesteps that minimizes cost terms~$C(\cdot)$ while also satisfying constraints formulated below,
\begin{small}
\begin{align*}
    \min_{\ddot{\theta}_{t\in{0,H-1}}} \quad & C_g(G, \theta_t) + C_{\text{sl}}(\theta_t, \dot{\theta}_t) \numberthis{eq:opt_cost}\\ &+  C_{\text{manip}} + C_{\text{stop}}\numberthis{eq:opt_cost_1},\\ 
    \text{s.t.} \quad &  S_e(\theta_t) < 0.0 \numberthis{eq:opt_env_coll}\\
    & S_r(\theta_t) < 0.0 \numberthis{eq:self_coll}\\
    & \dot{\theta}_t = \dot{\theta}_{t-1} + \ddot{\theta}_t dt \numberthis{eq:euler_vel}\\
    & \theta_t = \theta_{t-1} + \dot{\theta}_t dt \numberthis{eq:euler_pos}
\end{align*}
\end{small}
where~(\ref{eq:opt_cost}) contains the cost terms formulated for human-robot handover,  and~(\ref{eq:opt_cost_1}) lists the manipulability and stop cost terms from Bhardwaj~\etal~\cite{bhardwaj2021fast} that aid MPC in avoiding local minima and overshooting respectively~(refer~\cite{bhardwaj2021fast} for more details). Constraints~(\ref{eq:opt_env_coll}-\ref{eq:self_coll}) are collision avoidance constraints that prevent the robot from colliding with the environment, human, and itself. Eq.~(\ref{eq:euler_vel}-\ref{eq:euler_pos}) are euler integration equations to obtain joint position, and joint velocity from joint acceleration. We also have box constraints on the robot's joint position, velocity, and acceleration to satisfy joint limits.

\subsubsection{Reaching Grasp Pose}
Given a target pose~$X_g$ and the current gripper pose~$X_t=FK(\theta_t)$ computed using the forward kinematics of the robot at the current joint configuration~$\theta_t$, we define a pose distance metric~$\text{dist}(X_t, X_g)$,
\begin{align}
  \text{dist}(X_t,X_g) &= ||\alpha_1 (I - \T{R}{w}{g}^\top  \T{R}{w}{t})||_2 \nonumber \\ &
  +  || \alpha_2( \T{R}{w}{g}^\top \T{d}{w}{t} - \T{R}{w}{g}^\top \T{d}{w}{g}) ||_2, \label{eq:pose}
\end{align}
where~$\T{R}{w}{g}$ and~$\T{R}{w}{t}$ are the rotation matrices of pose~$X_g$ and~$X_t$ respectively. The translation vectors of~$X_g$ and $X_t$ are represented by~$\T{d}{w}{g}$ and~$\T{d}{w}{t}$ respectively.

We can reach a given gripper pose by using the above distance metric as the goal cost~$C_g(\cdot) = \text{dist}(\cdot)$. 
We can also optimize for reaching one pose from a set of grasp poses~$G$ by writing the goal cost as the distance between the current gripper pose~$X_t$ and the closest pose in the goal set, 
\begin{align*}
    C(\theta_t, G) &= \min(\text{dist}(FK(\theta_t), X_{g\in {G}})). \numberthis{eq:cost-goal}
\end{align*}

The above formulation of reaching a goal set enables MPC to reason about all the grasp poses inside the optimization while also considering all the other heuristics and constraints. This would enable MPC to move fluidly to another grasp from the current grasp given new constraints (\eg, human hand moves to current grasp region). We discuss results from both these formulations in Sec.~\ref{section:sys_eval}.

\subsubsection{Straight Line Cost}
We found the robot to take very circular trajectories in the Cartesian space due to the presence of many revolute joints in the manipulator. These circular trajectories are hard to predict by users, especially those with minimal domain knowledge. We hence formulate a cost term that penalizes gripper's linear velocity in directions that are not parallel to the vector connecting the current gripper position to the goal position~$\hat{d}_g=\frac{d_g - d_t}{||d_g - d_t||}$. We compute the gripper's current velocity leveraging the kinematic Jacobian~$J(\theta_t)$ at the current joint configuration~$\theta_t$ and the current joint velocity~$\dot{\theta}_t$ as~$\dot{d}_t=J(\theta_t)\dot{\theta}_t$ and then normalize this vector to get the current gripper motion direction~$\hat{d}_t$. The cost is then written as,
\begin{align*}
C(\hat{d}_t, \hat{d}_g) &= 1 - \hat{d}_t \cdot \hat{d}_g.
\numberthis{eq:sl}
\end{align*}
This cost becomes zero when both vectors are parallel.

\subsubsection{Collision Avoidance}
\begin{figure}
    \centering
    \includegraphics[width=0.49\textwidth]{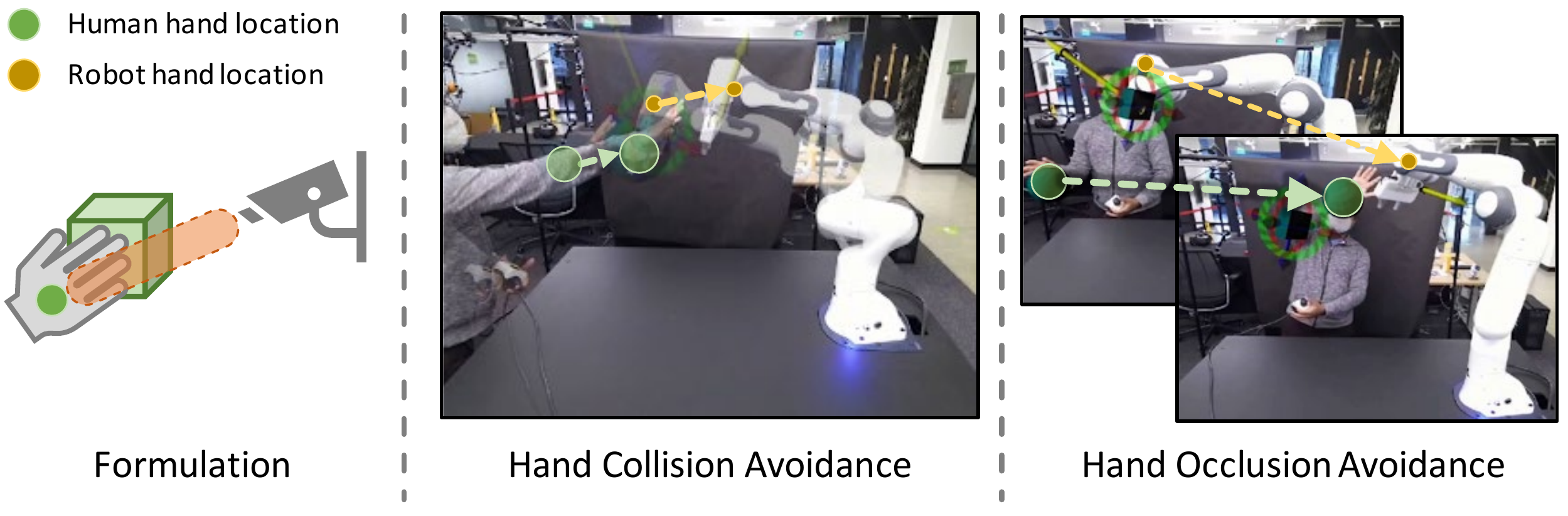}
    \caption{Left: the formulation of the hand collision and occlusion avoidance.
    The robot maintains an end-effector position marked by the green cube until the human hand either comes near the robot~(collision, middle) or gets occluded by the robot (right).}
    \label{fig:hand_avoidance}
    \vskip -18pt
\end{figure}

During gripper motion, the robot needs to avoid colliding with the table, and also the human hand. 
In addition, we want to keep the hand from being occluded by the robot during the robot motion. 
To do so, we represent the table as a cuboid, the human hand as a sphere, and use the line connecting the camera's origin with the human hand position to build a capsule as shown in Fig.~\ref{fig:hand_avoidance} (left). We also represent the robot's links with spheres similar to~\cite{bhardwaj2021fast} and then compute the signed distance between the robot and the environment via an analytic function~$S_e(\theta_t)$. We additionally avoid robot self collisions by using a neural network trained by~\cite{bhardwaj2021fast}. The neural network outputs the signed distance~\footnote{Our signed distance functions output negative values when there is no collision and positive when there is a collision.} between the two closest links given a joint configuration~$\theta_t$. We then use this as~$S_r(\theta_t)$ in~Eq.(\ref{eq:self_coll}). 
As demonstrated in Fig.~\ref{fig:hand_avoidance}, the robot avoids colliding with human hand and keeps the hand from occlusion.

\begin{figure*}[!t]
    \centering
    \setlength\tabcolsep{1.5pt}
	\begin{tabular}{ccc}
        \includegraphics[width=0.33\textwidth]{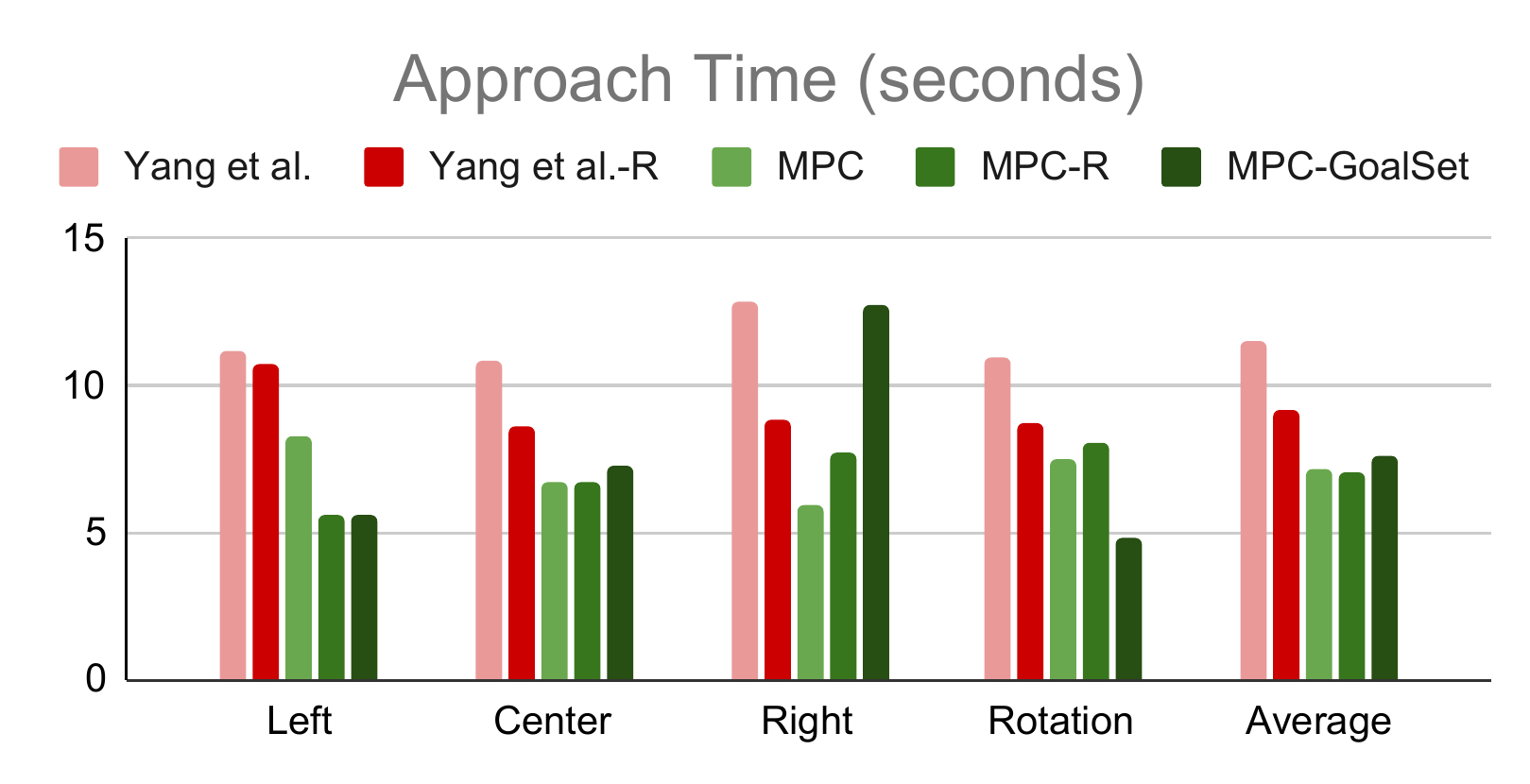} &
        \includegraphics[width=0.33\textwidth]{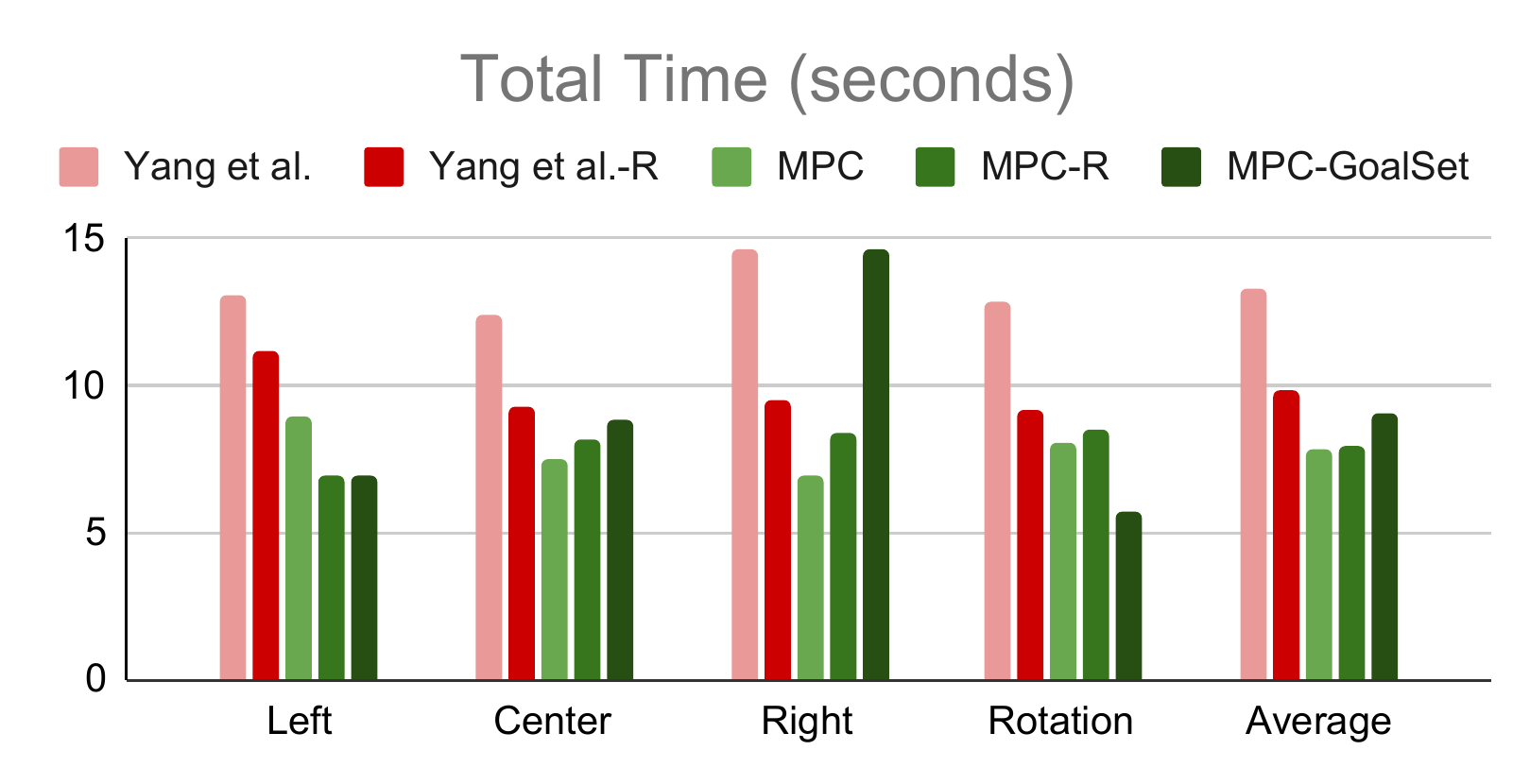} &
        \includegraphics[width=0.33\textwidth]{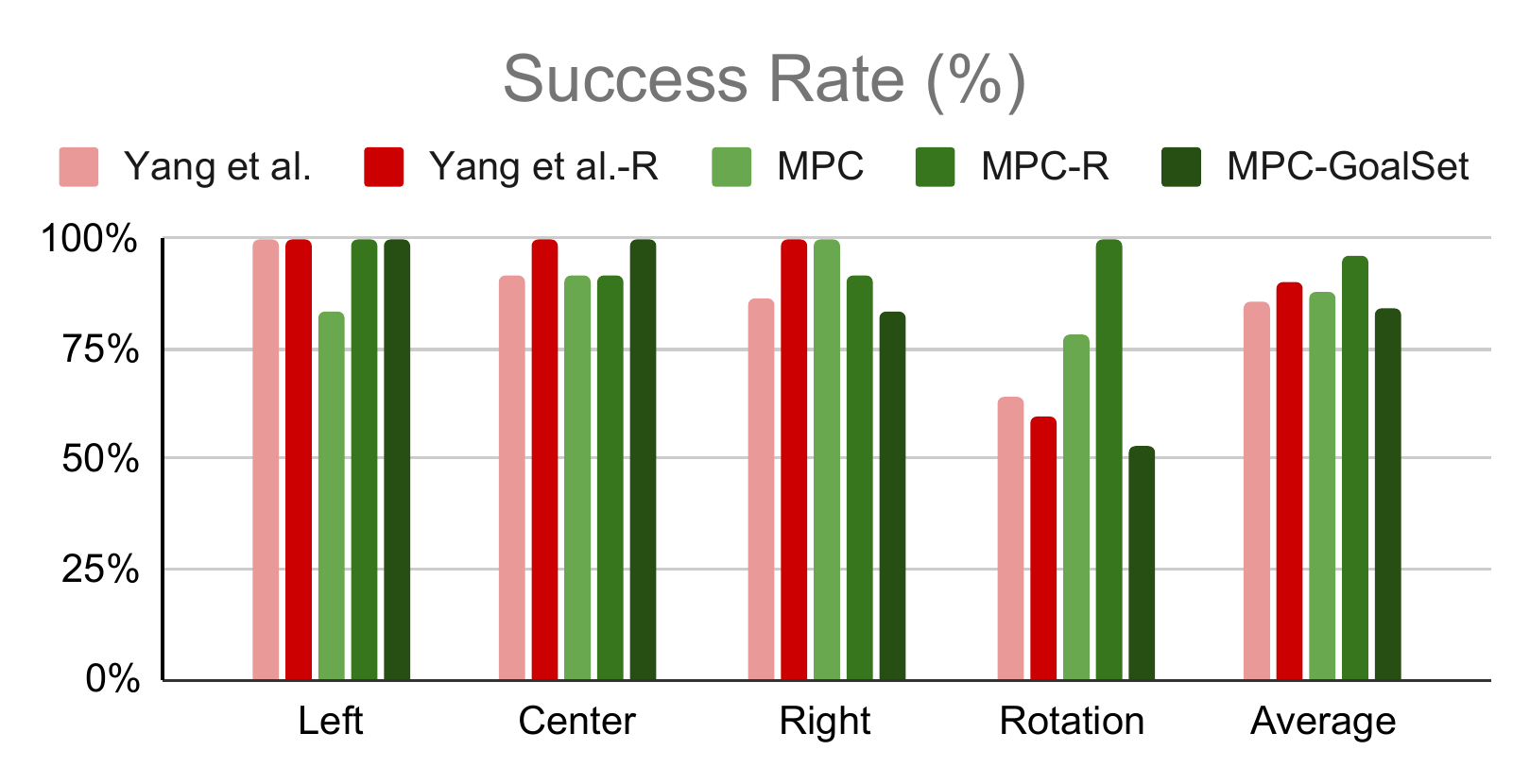} 
    \end{tabular}
  \vspace{-3mm}
  \caption{Systematic evaluation.
  *-R denotes methods with the reachability model.
  Our MPC framework generates faster and smoother motions compared with Yang \etal ~\cite{yang2021reactive}.
  }
  \label{fig:systematic-eval-location}
  \vspace{-12pt}
\end{figure*}

\subsection{Solving as Stochastic MPC}
\label{sec:mpc_solve}

We solve the optimization problem by leveraging stochastic MPC (also known as sampling-based MPC) as it has shown to work with many cost terms~\cite{bhardwaj2021fast}. Stochastic MPC optimizes by sampling action sequences for many particles from a distribution, rolling out these actions, computing the cost incurred by each particle, and then using these costs to update the distribution. By iterating through this process, given large number of particles, stochastic MPC can generate motions at 50-100 Hz which is comparable to gradient based MPC methods while having the benefit of not requiring differentiability of the cost terms or the dynamics~\cite{bhardwaj2021fast}. Stochastic MPC cannot handle constraints, hence we setup the constraints in our optimization problem as cost terms with large weights which has shown to be sufficient for satisfying constraints on realistic settings~\cite{bhardwaj2021fast,bhardwaj2021fast_exp}.
We also leverage this optimization problem for motion generation in other stages of the handover, \eg,  ``Drop", where we change the goal cost to be a $L2$ loss on the joint configuration. Our MPC uses a horizon of 2 seconds while optimizing the motion. All constraints have a weight of 5000. We set $\alpha_1$ and $\alpha_2$ in Eq.~(\ref{eq:pose}) as $70$ and $220$ respectively and $30$ for the straight line cost weight.
by starting with a small value and slowly increasing until the motion reached sufficient precision (a total of 30 minutes was spent to tune these parameters).

%% file: 7_systematic_evaluation.tex
\begin{figure}
    \centering
    \includegraphics[width=1\linewidth]{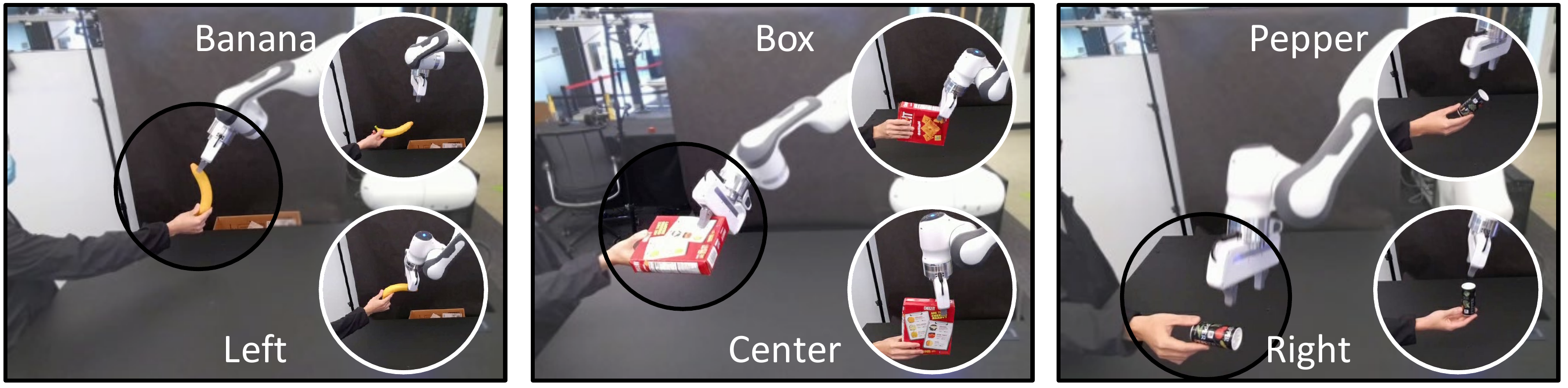}
    \caption{Systematic evaluation on different handover locations. For each location, we handed over the object three times with three different ways of holding the object.
    }
    \vskip -18pt
    \label{fig:handover-location}
\end{figure}

\smalltitle{System Setup.} 
We conducted experiments with a Franka Emika Panda arm, and an externally mounted Azure Kinect RGBD camera with $1280\times 720$ resolution. 
The system is distributed on 3 desktop computers with an additional realtime desktop for Franka control. Four NVIDIA RTX 2080 Ti and one 3090 GPUs were used.

\smalltitle{Protocols and Metrics.} 
We evaluated the system performance with three objects with different shapes and sizes, including a banana from the YCB object~\cite{calli2015ycb}, a cracker box, and a pepper bottle, as shown in Fig.~\ref{fig:handover-location}. 
For all the experiments in this section, we repeated the handover trials until three successful handovers have been recorded. 
Then we measured the system performance by computing the \textit{success rate,  approach time}, and the \textit{total time} for a successful handover. 
We also recorded the \textit{velocity, acceleration} and the \textit{jerk} during the handover.

\smalltitle{Baselines.} 
We considered Yang \etal~\cite{yang2021reactive} as the baseline method. In addition to the grasp selection criterion proposed in~\cite{yang2021reactive}, we use the learned reachability metric $R(X_{g})$ described in Sec.~\ref{sec:reachability} to determine which grasps are reachable and restrict the set to ones that are highly manipulable:
\begin{align}
    C = &\omega_{s} \min(S_{g} - S_{min}, 0)
    + \omega_{prev} d(X_{g}, X_{prev}) \nonumber\\
    &+ \omega_{home} d(X_{g}, X_{home}),
    + \omega_{R} R(X_{g}),
\end{align}
where $S_{g}$ and $S_{min}$ is the score for grasp $g$ and the minimum acceptable score, $X_{g}$ and $X_{prev}$ denote the pose of grasp $g$ and the pose of previous chosen grasp respectively. 
$X_{home}$ denotes the end-effector pose at the home position. $\omega_{s}, \omega_{prev}, \omega_{home}, \omega_{R}$ are weights. 
$d(\cdot, \cdot)$ is a distance metric with both position and rotation components. 

We compared the above baselines with the following variations of the proposed approach: 1) \textit{MPC/MPC-R}: using MPC for motion planning towards one selected grasp (without/with reachability metric); 2) \textit{MPC-GoalSet}: using MPC to generate motion given a set of grasps. Noted that reachability metric is inherently embedded in the cost function Eq.(\ref{eq:opt_cost_1}).

\subsection{Handover Reactivity}

We first investigate the performance of our approach by rotating the object along it's standing axis 45 degrees after the robots starts moving. 
As shown in Fig.~\ref{fig:systematic-eval-location} (Rotation), all \textit{MPC}-based systems have lower approach/total time for successful handovers since they can react quickly to the changing of the object orientation. 
The reachability model promotes lower approach time while maintaining a similar success rate for the baseline method~\cite{yang2021reactive}, and improves the success rate for the \textit{MPC} variant. 
We observe that while the \textit{MPC-GoalSet} achieves the lowest approach time for successful handovers, it has a reduced success rate especially when handling changes in orientation. 
We suspect this to be because the number of grasps sufficiently reduces after the filtering by the reachability model (~Sec.~\ref{sec:reachability}), and at least one grasp is consistent across time.

\subsection{Handover Location}

\begin{figure*}
    \centering
    \includegraphics[width=0.95\linewidth]{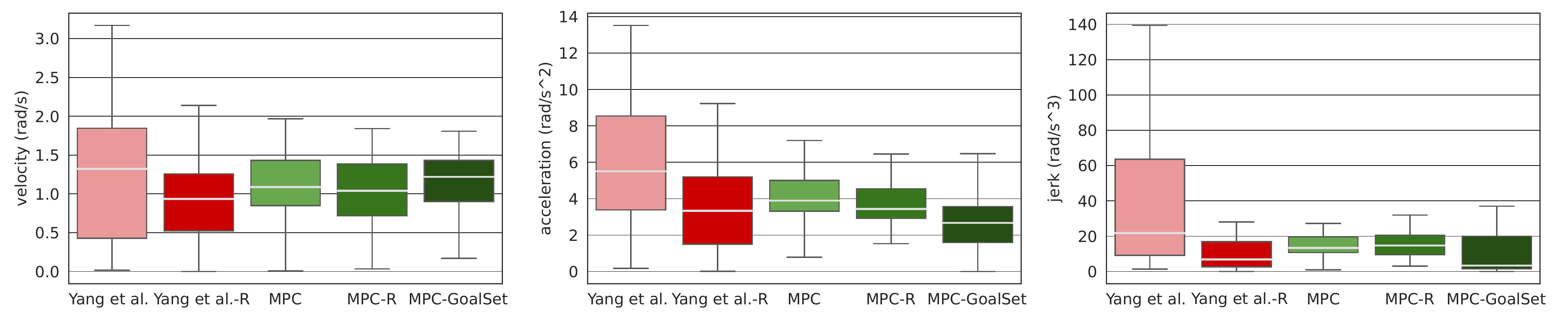}
    \vspace{-2pt}
    \caption{Our proposed approach reduces jerk compared with prior work~\cite{yang2021reactive}, a possible explanation for the better user experience.
    }
    \label{fig:motion_stats}
\end{figure*}
We further evaluate the efficiency of our approach in response to H2R handovers at different locations and different ways of holding objects.
We handed over three objects at three locations: \textit{left}, \textit{center}, and \textit{right}. 
For each location, we handed over the object three times with three ways of holding the object, as shown in Fig.~\ref{fig:handover-location}. 
Results are reported in Fig.~\ref{fig:systematic-eval-location}.

The \textit{MPC}  is comparable to Yang \etal~\cite{yang2021reactive}, where we used \textit{MPC} for motion planning while keeping the robot velocity (Fig.~\ref{fig:motion_stats} left) and all the other modules the same. 
\textit{MPC} reduces the approach time ($11.5$s to $7.1$s) and total time ($13.3$s to $7.9$s)  and improve the success rate from $85.7\%$ to $88.3\%$.

We also investigate integrating the reachability model with both Yang \etal~\cite{yang2021reactive} and \textit{MPC} (denoted as \textit{*-R}). 
In this setting, the reachability metric allows for better grasp selection. 
It improves the approach time of~\cite{yang2021reactive} by two seconds and also improves the success rate of MPC by $7.5\%$.
This is because, with the reachability model, the robot tends to choose more reachable/manipulable grasp poses. 
Our \textit{MPC-GoalSet} variant has slightly longer approach time and slightly lower success rate compared to \textit{MPC/MPC-R}, but it achieves smoother motion with less jerk (see Fig.~\ref{fig:motion_stats} and Sec.~\ref{sec:motion-metrics}).

\begin{table*}[]
\centering
\setlength\tabcolsep{1.8pt}
\begin{tabular}{llccccccccccc}
\toprule
&       & box    & newspaper & plate & mug & remote & toothpaste & scissors & towel & pen & spoon & average \\
\midrule
\multirow{2}{*}{Appr. Time} & 
Yang \etal~\cite{yang2021reactive} &  7.5$\pm$2.4	& 8.8$\pm$1.0 & 10.0$\pm$1.2 & 14.0$\pm$6.3 & 11.5$\pm$6.6 & 9.0$\pm$1.2 & 10.5$\pm$5.8 & 9.5$\pm$3.1 & 9.5$\pm$1.3 & 10.5$\pm$1.0 & 10.1$\pm$3.7  \\
 & Ours   & 6.5$\pm$2.5 & 6.3$\pm$1.0	 & 5.8$\pm$2.1 & 5.3$\pm$1.3	 & 8.8$\pm$7.1 & 10.8$\pm$6.7 & 15.3$\pm$4.9 & 4.3$\pm$1.3 & 6.5$\pm$3.9 & 5.8$\pm$2.1 & 7.5$\pm$4.7  \\
\midrule
\multirow{2}{*}{Success Rate} 
& Yang \etal~\cite{yang2021reactive} & 66.7\% &	100.0\%	& 66.7\% & 	100.0\% &	80.0\% & 	66.7\% &	66.7\% &	80.0\% &	50.0\%	& 80.0\% &	75.7\%  \\
& Ours  & 66.7\%	& 80.0\% &	100.0\%	& 66.7\% & 	50.0\% & 	66.7\% &	80.0\%	& 66.7\%	& 100.0\%  &	66.7\% &	74.3\%
\\ \bottomrule
\end{tabular}
    \caption{Quantitative results from the user evaluation with four users.
    }
    \vspace{-2em}
    \label{tab:user_study}
\end{table*}

\subsection{Motion Metrics}\label{sec:motion-metrics}
To better understand the motion pattern, we recorded the robot position, velocity, acceleration and jerk during the above experiments. 
The statistics are reported in Fig.~\ref{fig:motion_stats}. 
In general, all our MPC-based approaches have more consistent velocity and accelerations compared with the baseline approach~\cite{yang2021reactive}. 
Our approach has less sudden accelerations, resulting in less jerk. 
In particular, our MPC-GoalSet achieves the least jerk, as in this formulation MPC optimizes to reach a single grasp from the grasp set, while considering the robot's current position, velocity, and acceleration.

%% file: 8_user_study.tex
We also conducted a user evaluation to compare the proposed system (\textit{MPC-R} as \textit{Ours}) with the prior system Yang \etal~\cite{yang2021reactive} to validate that
our system enables fluid H2R handovers. 
We recruited four participants from the lab\footnote{We were not able to recruit outside participants due to the ongoing COVID-19 pandemic.}. 
Each participant did two rounds of handovers to interact with two systems without knowing the details of the systems. 
Participants were instructed to hand over ten objects from Household-A objects used in~\cite{yang2021reactive} to the robot one at a time. 
They were asked to rate the systems with a set of Likert scale questions that are commonly asked in prior works~\cite{ortenzi2020object} immediately after each round of interaction with the system. 
We additionally asked the users which system is less jerky to evaluate the subjective motion smoothness. 
We counterbalanced the order in which participants interacted with the two systems to avoid the bias.
After completing all interactions, participants were asked to rank the two systems in different dimensions and share their opinions and comments through open-ended questions. 

\smalltitle{Objective Evaluation.}
We report the approach time and the success rate in Table.~\ref{tab:user_study}. 
Our approach was able to take over the object with less time except for \textit{scissors} and \textit{toothpaste}, and the overall approach time is shorter than the prior system~\cite{yang2021reactive}. 
The success rate of our system is on par with prior system~\cite{yang2021reactive}.

\begin{figure}[t]
    \centering
    \includegraphics[width=1\linewidth]{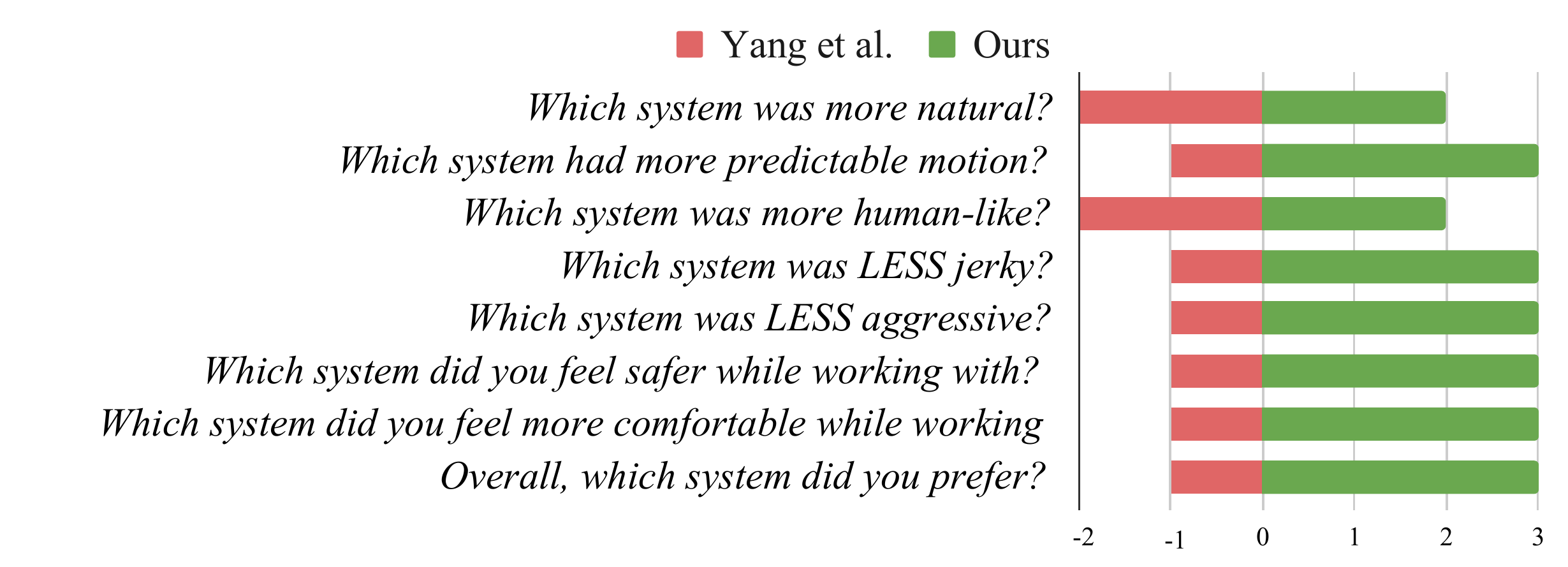}
    \caption{Participants' ranking of the two systems across eight questions in the questionnaire.
    Length of the bar denotes the number of participants. 
    }
    \label{fig:handover-rating}
    \vskip -0.5cm
\end{figure}

\smalltitle{Subjective Evaluation.}
Fig.~\ref{fig:handover-rating} reports user ratings over the two systems. 3 out of 4 users preferred our proposed system. 
Particularly, they thought our system was more predictable, less jerky, less aggressive, safer, and more comfortable to work with. 
The main reasons for choosing our systems are ``I was able to predict the robot motion better", ``it felt less aggressive", and ``it did not crash" and ``required less motion before picking the object". 
One user preferred~\cite{yang2021reactive}, commenting ``(\cite{yang2021reactive}) had a better approach for some of the items and changed trajectories for less items, thus providing a better experience for me". 
But this same user also commented ``some of these questions are really A=B", which proves the user experience of our system is better or on par with the prior system~\cite{yang2021reactive} in these aspects. 
For questions ``which system was more natural" and ``human-like", half of the users chose~\cite{yang2021reactive} and the other half preferred our system. 

%% file: 9_conclusion.tex
\begin{figure*}[htp]
    \centering
    \includegraphics[width=0.7\linewidth]{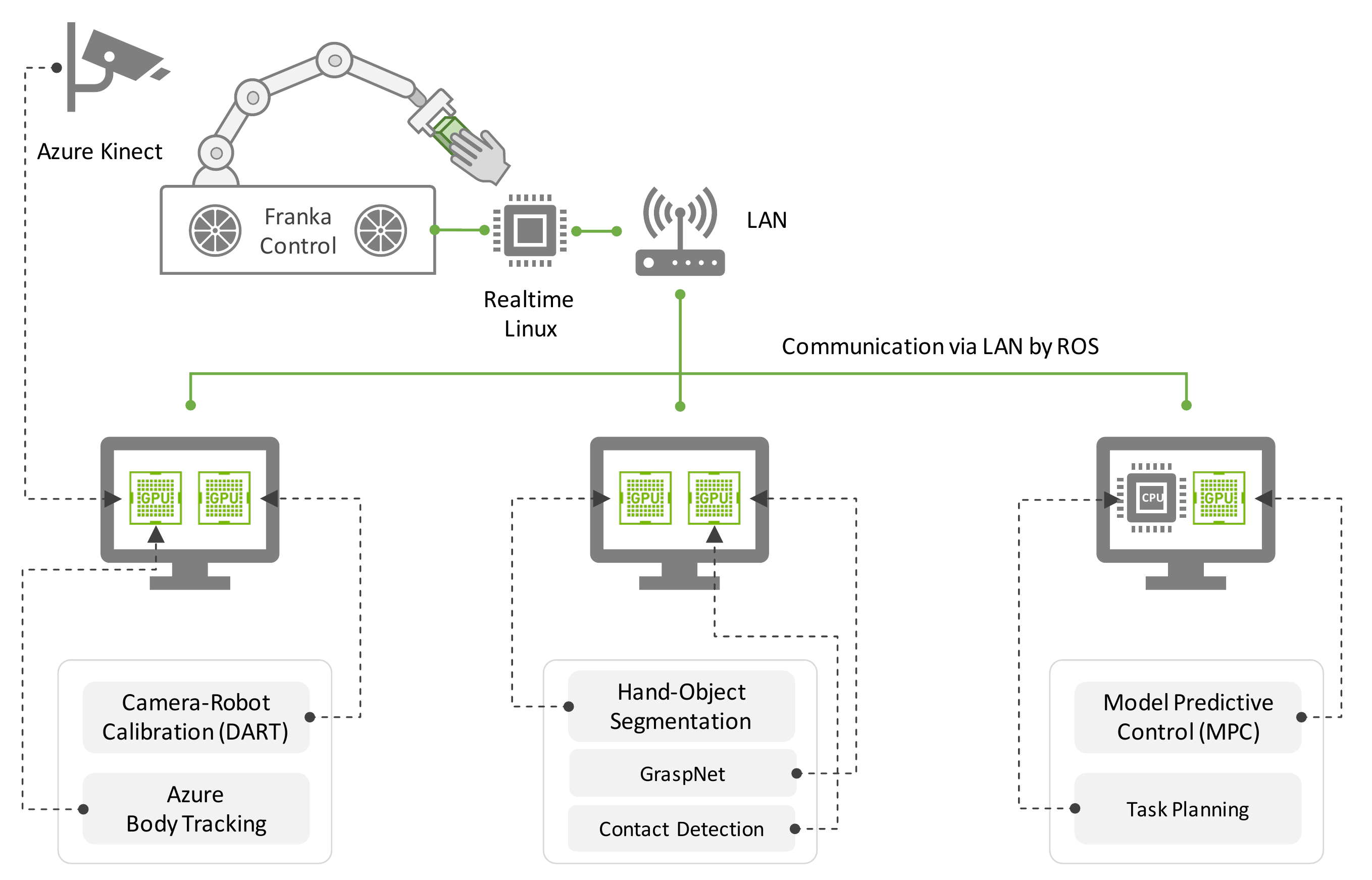}
    \caption{The structure of the proposed handover system.}
    \label{fig:system-diagram}
\end{figure*}

We proposed a method for fluid human-to-robot handovers by incorporating more knowledge about the robot motion planning, specifically using MPC to find smooth, consistent trajectories. 
To facilitate grasp selection, we learned a reachability model to prioritize robot grasps with higher manipulability scores. 
Our approach was demonstrated more efficient through a systematic evaluation and more favorable by users through a user evaluation compared to prior work.
Nevertheless, the reachability model and the grasp generation are independent, which sometimes leads to robot oscillations when the reachability scores and the candidate grasps are in disagreement, especially when using  MPC given a set of grasps (MPC-GoalSet).
In the future, we plan to better incorporate grasp and motion planning as well as reachability modeling to improve the robustness of the system.

%% file: 10_appendix.tex
\appendix
\subsection{System Diagram}\label{sec:system-diagram}
Fig.~\ref{fig:system-diagram} illustrates the structure of our system. 
We used a Franka Emika Panda arm for all experiments.
The Franka control box was connected to a realtime Ubuntu desktop. 
An Azure Kinect RGBD camera was mounted next to the robot to capture the full scene. 
The perception, planning and control modules are distributed on three desktop computers.
Specifically, the Azure Kinect software, including depth image processing and body tracking, and the camera-robot calibration using DART~\cite{schmidt2014dart} ran on two NVIDIA RTX 2080 Ti GPUs from the same desktop.
The hand-object segmentation, GraspNet~\cite{mousavian2019graspnet}, and the physical contact detection net ran on another two NVIDIA RTX 2080 Ti GPUs from a different desktop.
Finally, the stochastic MPC is computed from an NVIDIA RTX 3090 GPU and the task planning is computed without GPUs.
All the desktops are connected through a local area network (LAN) and communicated through the Robot Operating System (ROS).

\subsection{Neural Network Architectures}
\subsubsection{Physical Contact Detection Model}
\begin{figure}[tp]
    \centering
    \includegraphics[width=1\linewidth]{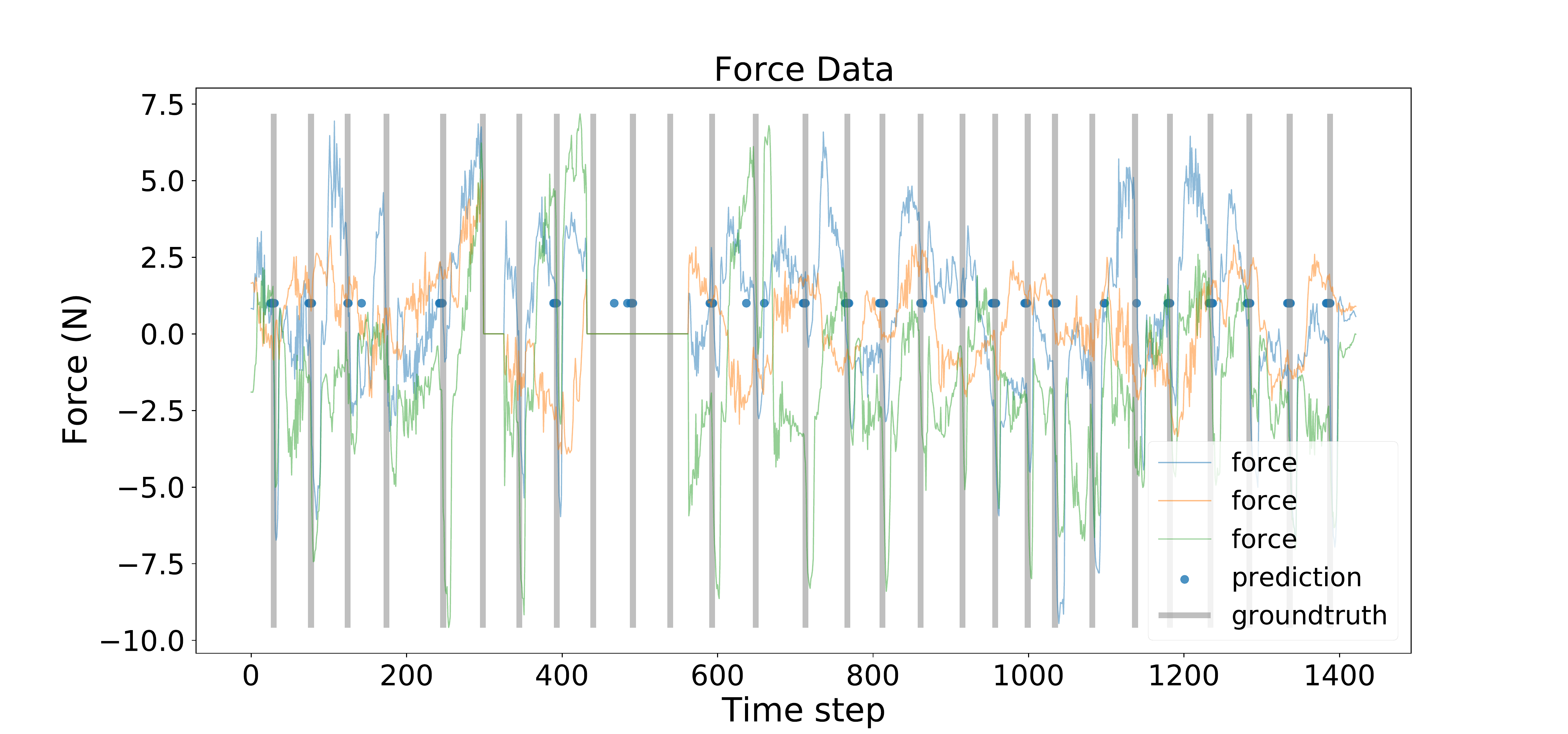}
    \caption{An example of the captured force data. Gray vertical lines denote the moment when a contact occurs. Blue circles illustrate the detected contact by our model. }
    \label{fig:force}
\end{figure}

Our contact detection model takes a concatenation of joint velocities (7 dim), efforts (7 dim), force (3 dim) and torque (3 dim) for the past $T=5$ steps as input, which is a $T \times 20$ tensor. 
Each individual timestep is first encoded through a two-layer perception.
Each layer consists of a 1D convolutions with kernel size one and a ReLU activation (the number of channels is given by 20–256,256–512). 
Then the model performs two 1D convolutions with kernel size one and 512 channels with ReLU activation and Batch Normalization over history. 
Finally, another two-layer perception (1D convolutions with kernel size 1 and number of channels as 512-256, 256-1, ReLU activation and dropout) predicts a single output indicating if a force event was detected within the window. 
The model is trained with the binary cross-entropy loss. 
Fig.~\ref{fig:force} visualizes the recorded force, the ground truth label and our prediction of a test sequence.

\subsubsection{Reachability Model}
The reachability prediction model takes in the 6-dim grasp pose as input, which is passed through five fully-connected layers, each with 64 neurons and ReLU activation. The final output is the predicted reachability value and the model is trained with the mean-squared-error loss.

\begin{figure*}[ht]
    \centering
    \begin{tabular}{c c c c c c}
         \includegraphics[width=0.14\linewidth]{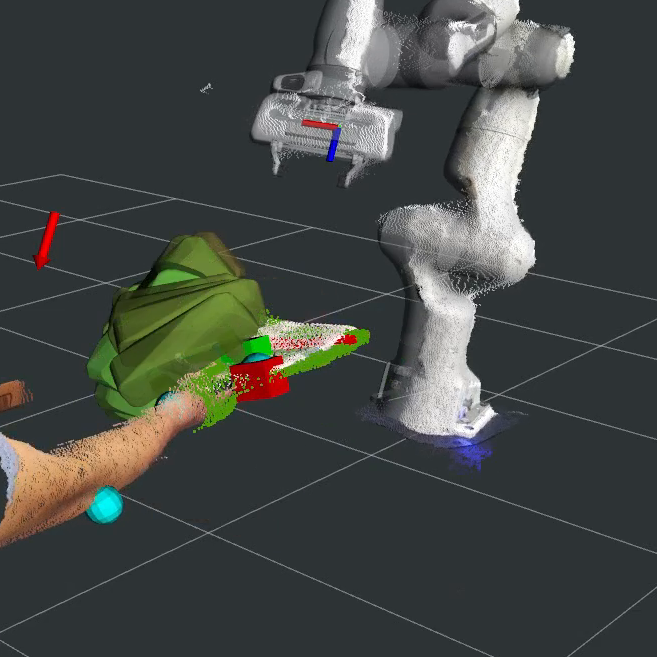} & 
         \includegraphics[width=0.14\linewidth]{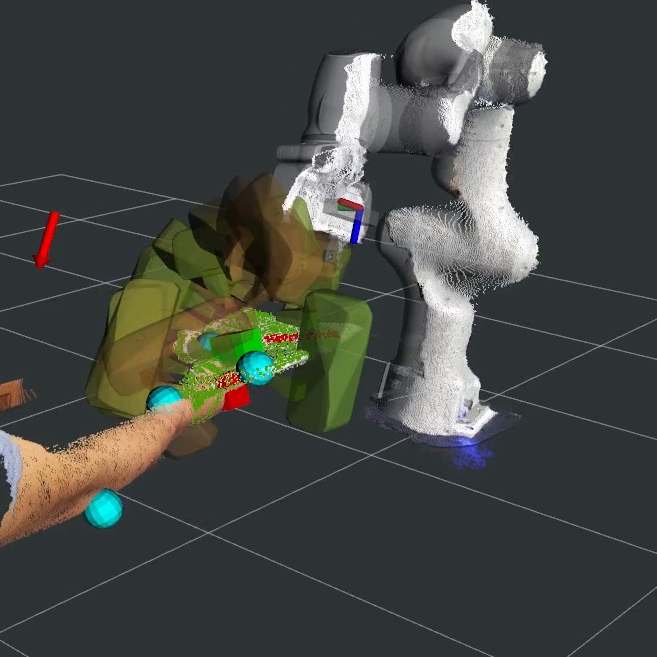} & 
         \includegraphics[width=0.14\linewidth]{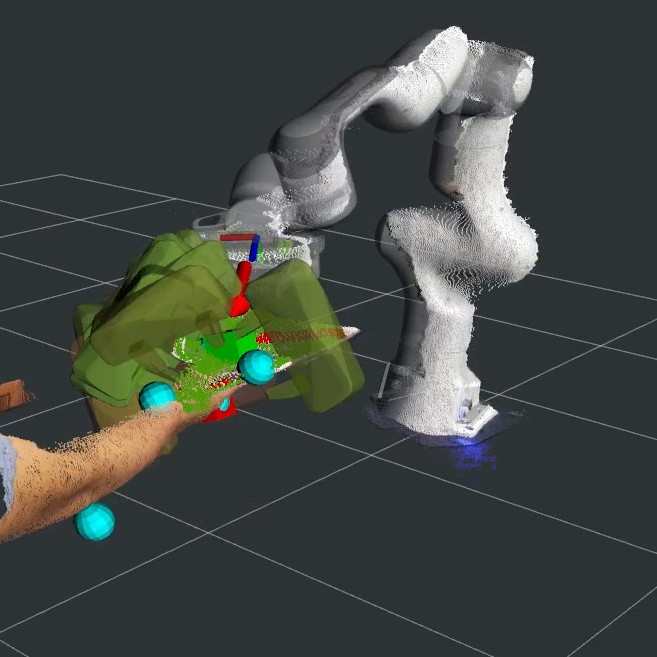} &  
         \includegraphics[width=0.14\linewidth]{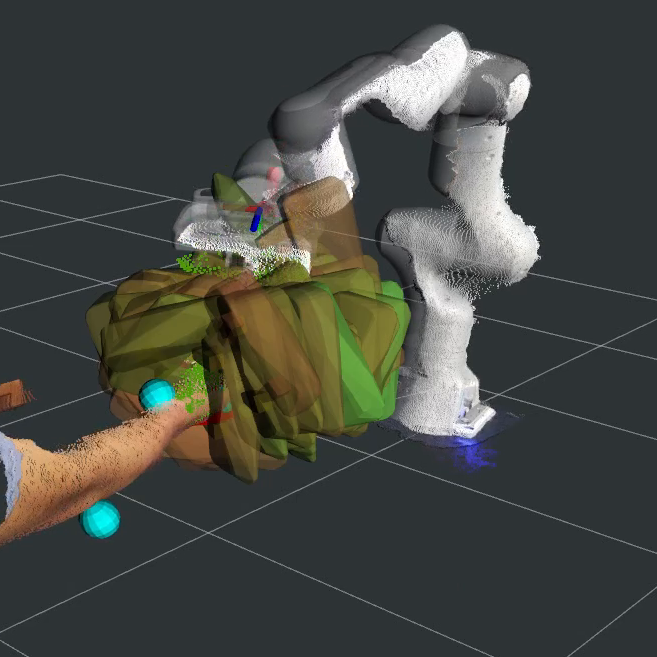} & 
         \includegraphics[width=0.14\linewidth]{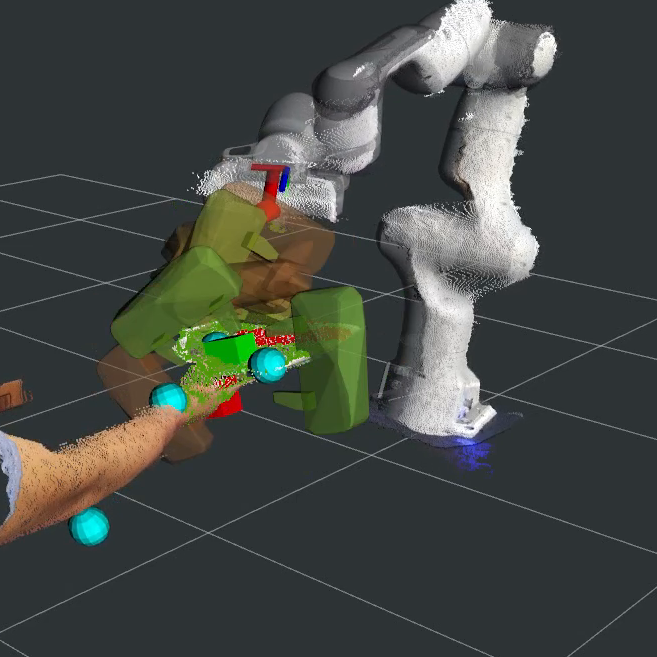} &
         \includegraphics[width=0.14\linewidth]{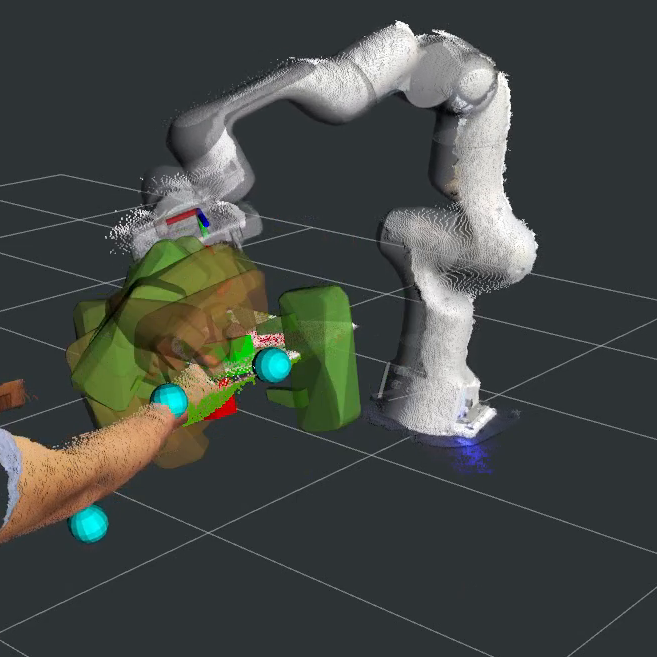} \\
         $T=17$ & $T=25$ & $T=38$ & $T=48$ & $T=60$ & $T=80$    
         \end{tabular}
    \caption{The oscillation of robot motion caused by inconsistent grasps.}
    \label{fig:inconsistent-grasp}
\end{figure*}
\subsubsection{Hand Segmentation Model}
The backbone of the hand segmentation model is a Feature Pyramid Network~\cite{lin2017feature} based on ResNet-50~\cite{he2016deep} pretrained on the ImageNet~\cite{russakovsky2015imagenet}. The four convolutions with upsampling operations are used to recover the feature map with the original resolution and a binary segmentation mask is predicted to indicate whether a pixel is hand or background. Please see~\cite{yang2021reactive} for details of the training dataset.

\subsection{Failure cases}
Graspnet~\cite{mousavian2019graspnet} and reachability together do not guarantee consistent grasps. As a result, we do see a number of cases where no grasps are possible -- for example, GraspNet might predict only grasps that the reachability model says are very low quality. 
One problematic case sees a number of generated grasps that are right on the threshold. 
Due to the noise from one step to the next, the method might be trying to reach very different results, leading to oscillations that are unnerving to a human user, as shown in Fig.~\ref{fig:inconsistent-grasp}.